# Online Social Support Detection in Spanish Social Media Texts

Moein Shahiki Tash[a], Luis Ramos[a], Zahra Ahani[a], Raúl Monroy[b], Olga kolesnikova[a], Hiram Calvo[a], Grigori Sidorov[a]

[a]*Instituto Politécnico Nacional (IPN), Centro de Investigación en Computación (CIC), Mexico City, Mexico,*
[b]*Tecnologico de Monterrey Escuela de Ingenieria y Ciencias*

**Abstract**

The advent of social media has transformed communication, enabling individuals to share their experiences, seek support, and participate in diverse discussions. While extensive research has focused on identifying harmful content like hate speech, the recognition and promotion of positive and supportive interactions remain largely unexplored. This study proposes an innovative approach to detecting online social support in Spanish-language social media texts. We introduce the first annotated dataset specifically created for this task, comprising 3,189 YouTube comments classified as supportive or non-supportive. To address data imbalance, we employed GPT-4o to generate paraphrased comments and create a balanced dataset. We then evaluated social support classification using traditional machine learning models, deep learning architectures, and transformer-based models, including GPT-4o, but only on the unbalanced dataset. Subsequently, we utilized a transformer model to compare the performance between the balanced and unbalanced datasets. Our findings indicate that the balanced dataset yielded improved results for Task 2 (Individual and Group) and Task 3 (Nation, Other, LGBTQ, Black Community, Women, Religion), whereas GPT-4o performed best for Task 1 (Social Support and Non-Support). This study highlights the significance of fostering a supportive online environment and lays the groundwork for future research in automated social support detection.

## 1. Introduction

Social support is usually conceptualized as an emotional, intangible and tangible aid procured from one's social connections, whereby the person feels loved, cared for, respected, and valued. A plethora of research has established the idea that social support is beneficial and greatly enhances both psychological and physical well-being, acting as an important factor of protection (Xia et al., 2012), significantly reducing the risk of mortality, with an impact comparable to factors such as obesity or inactivity (Kent de Grey et al., 2018).

Social support is positively related to psychological and physical health (Bellinia et al., 2019). Recent studies rightly point out that its functions are vast especially as highly developed social networks can reduce risks and help solve major medical problems (Kent de Grey et al., 2018). Additionally, social support has been associated with fewer symptoms of depression, anxiety, and stress (Baeza-Rivera et al., 2022).

For the last few years, social media has rapidly grown as an aid to help establish good relations and dialogue among family, friends and other acquaintances. Nonetheless, this growth has not been without some disadvantages. Its unregulated nature sometimes causes online environment that is hostile. Frequently, they are used for spreading hate speech and posting offensive comments which have varied negative effects on the society (Abdelsamie et al., 2024). Thus, it is necessary to design strategies that will help notice such kinds of content with a view for creating safer digital spaces as well as promoting decent conduct over the internet.

Despite the impact it can have on people's well-being, the promotion of positive and supportive content has not received much attention in this area of research. In response to these challenges, our proposed approach offers a disjunct but underexplored strategy for combating negativity on social media platforms by promoting support comments. Instead of focusing efforts on detecting and filtering negative content, our approach seeks to cultivate a more positive and supportive online environment by encouraging users to provide emotional comfort, encouragement, and advice to those facing challenges.

Online social support refers to the emotional aid and consolation given via digital platforms such as social media. This kind of support is very important for individuals and groups that interact on social networks and face different challenges like victims of war, marginalized communities, minority populations etc. Digital platforms make it possible for users to connect with others who share similar experiences. That way they can get necessary resources, help, empathy, courage or encouragement they need. Online support networks are usually vital because they allow for anonymity and accessibility to people who do not have access to conventional supportive systems. They also enable real time participation which encourages a sense of belongingness, hence fostering psychological well-being by minimizing negativity effects towards the user.

The current study is based on the influence of social support on individual's health. YouTube was utilized to collect data starting from the search for some potential videos that could generate comments from various kinds of support. As such, 3,189 comments were collected in Spanish, of which only 679 express support, while 2,510 do not. Spanish is one of the most widely spoken languages globally, yet it remains underrepresented in NLP research. Analyzing social support in Spanish allows us to address linguistic and cultural differences, improve accessibility, and contribute to the development of multilingual social support detection models. Additionally, in this study, we used traditional machine learning models such as logistic regression (LR), support vector machine with radial-based function core (SVM(rbf)), vector support with linear core (SVM(linear)), XGBoost and Random Forest as baseline models, while deep learning models (BiLSTM and CNN) was tested using different word embeddings (fastText and GloVe) and finally, some transformers models and a LLM (GPT4-o) were tested. Additionally, tests were performed with a balanced dataset, for this over-sampling technique was used, so this study revealed that the best performance for Subtask 1 was achieved by GPT4-o using the original dataset, reaching a Macro F1 Score of 0.8531, while for Subtask 2 and Subtask 3 the transformel model "obertuito-sentiment-analysis" achieved better performance using the balanced dataset, reaching a Macro F1 Score of 0.8894 and 0.8361, respectively.

---

*These authors contributed equally to this work.
[1]Corresponding author: (mshahikit2022@cic.ipn.mx)

The main contributions of this research paper are listed below:

- Studying social support for social wellness as a novel task in NLP,

- Developing annotation guidelines and creating the first specific social support detection dataset in Spanish,

- Conducting extensive benchmark experiments using traditional machine learning, deep learning, transformer models, and GPT-4o,

- Performing balanced dataset experiments using the GPT model by paraphrasing comments,

- Comparing results obtained from balanced and unbalanced datasets.

## 2. Definitions

(Xia et al., 2012) expose a definition of social support as mental and material support obtained from the social network, making one feel that he is cared for, loved, valued and appreciated. The definition of social support has broadened to include the virtual assistance and connections that individuals form online, commonly referred to as online social support (Ma et al., 2024), that involves behaviours, communication, and interactions that demonstrate care and appreciation for individuals, thus fostering a sense of belonging and helping to cope with life's challenges (Kolesnikova et al., 2025).

This study aligns with the previous definition of social support and explores the exchange of comments between users and audiences as a form of social support occurring within social media.

## 3. Related work

Given the importance of promoting positive and supportive discourse, research on this topic remains relatively scarce. This concept serves as a counterpart to hate speech. However, there is no directly comparable research specifically focused on the Spanish-language support speech in social media using NLP techniques.

Social support has been studied recently using diverse NLP techniques. (Ahani et al., 2024) proposed the detection of supportive speech on social media using NLP techniques and ML and DL models. They demonstrate that the integration of psycholinguistic, emotional and feeling characteristics with n-grams can detect social support, and it is also possible to distinguish whether it is directed at an individual or a group. The best results obtained for the different binary and multiclass tasks in all experiments range from 0.72 to 0.82. (Kolesnikova et al., 2025) also proposed the use of LLM, on the same dataset as in previous research, applying Zero-Shot learning to models such as GPT-3, GPT-4, and GPT-4-turbo, Transformer models available on Hugging Face's website were also used, with RoBERTa-base consistently outperforming others by improving previous metrics by up to 8%.

(Zou et al., 2024) offer a comprehensive description regarding the self-disclosure processes and social support skills on online platforms, focusing on women suffering from infertility on Reddit. Their study combines several theories, such as Communication Privacy Management Theory, Functional Theory of Self Disclsoure, Social Support Theory, and Social Penetration Theory. These theories assist in assessing the consequences of personal disclosures on the amount and form of assistance offered in social media. Employing NLP techniques, the authors evaluate a corpus of Reddit posts and comments spanning across three years' worth of data. Precise text classification regarding self-disclosure and social support types is performed utilizing the BERT model. The model is shown to be effective with self-disclosure and social support through precision, recall, and F1 scores.

(Erčulj et al., 2019) proposed a text-mining approach to detect automatically discussion topics in the largest infertility forum in Slovenia and identify themes of social support types among patients coping with infertility. The study focused on an infertility forum where 13,2374 posts were made between 2002 and 2016. Topics of discussion were identified through the LDA method. The findings are suggestive that online forums, like health-related online groups, can offer critical support for patients with infertility issues, thus confirming hypotheses from earlier research. The results illustrate the efficacy of text-mining to understand and analyse online social support behaviour and help in communication within the health care system. Later, (Erčulj and Pavšič Mrevlje, 2023) tested LDA, but in this study it was to analyze the women in need. The study suggests that increasing user engagement and possibly integrating more structured support mechanisms could enhance the effectiveness of online support communities for women in need.

## 4. Dataset development

### 4.1. Data collection and processing

Our study analyzes a corpus of YouTube comments aimed at Spanish-speaking audiences on videos covering diverse topics, including but not limited to, nationality, the Black community, women, religion, and LGBTQ+ issues. The videos chosen were those that garnered significant support, such as content related to the Olympic Games or those focused on social issues related to race, gender, and sexual orientation. The dataset consists of 3,189 comments, having been cleaned of duplicate comments and those not in Spanish.

In this regard, it must be pointed out that no other filtering or selection process was performed over the comments associated with the videos selected. This strategy made it possible to analyse issues pertaining to the articulation of support, but still prevailing in the comments focused on the actual quantitative distribution in the videos.

### 4.2. Annotator selection

For the selection of annotators, three male native Spanish-speaking candidates were recruited, two of whom were pursuing master's degrees in computer science. To ensure consistency and accuracy in the annotation process, each annotator was initially provided with a set of 100 sample tweets along with a comprehensive guide outlining the annotation protocols. This allowed the annotators to become familiar with the task before data generation and collection began.

Following this, the labeled samples from the first two annotators were thoroughly reviewed and analyzed. To address any discrepancies or challenges, individual meetings and interviews were held with each annotator. During these discussions, the annotators provided insights into the issues and conflicts they encountered while labeling the data, which helped refine the annotation process.

Recognizing the need for a highly experienced annotator to maintain the quality of the annotation, the authors selected a third annotator. This third annotator was one of the authors of the paper, a PhD student specializing in Natural Language Processing. As a native Spanish speaker with extensive knowledge of the subject matter, she contributed her expertise to ensure the consistency and accuracy of the annotation process. Her involvement played a crucial role in finalizing the labeled dataset.

### 4.3. Annotation guidelines

The Social Support detection task was structured as a three-step classification process. First, supportive comments were identified. Next, it was determined whether these supportive comments were directed toward an individual, a group, or a community. Finally, if the supportive comment was identified as being directed toward a group, the specific group was further identified. The guidelines for this process are described below.

- **Subtask 1 - Binary social support detection:** In this subtask, a given text is classified as supportive or non-supportive:
    - **Social Support (label = SS):** Statements of support promote understanding, empathy and positive actions. Therefore, a supportive comment is a statement or message that offers support, encouragement, admiration, advocacy, promotion, assistance, or defence. These comments are intended to provide emotional support, raise morale, recognize the achievements, or labour of others.
    - **Not Social Support (label = NSS):** The text does not convey any form of support as specified by the previous definition.

- **Subtask 2 - Individual vs. Group:** In this subtask, each supportive comment identified in the previouse subtask is categorized as either individual support or group support.
    - **Individual:** If the text expresses support for a specific person or individual (e.g., Alan Turing, Nikola Tesla, Donald Trump, Steve Jobs, etc.), it is labelled as Individual.
    - **Group:** If the text expresses support for a group of people, community, nation, etc. (e.g., Christians, Black community, LGBTQ or Other), it is labelled as Group.

- **Subtask 3 - Multiclass SS for Groups:** In this subtask, the aim is to identify which community or group of people receive social support by classifying the group support comments identified in subtask 2 into the following categories:
    - **Nation:** This category includes texts that express support for a specific country, its people, or its sovereignty. Examples might include advocating for the rights of a nation, showing solidarity upon the occurrence of a national crisis, or celebrating a country's achievements. For instance, "Sending love and strength to Ukraine during these challenging times" would fall under this category.
    - **Black community:** This type of support acknowledges and uplifts the Black community, often focusing on racial equality, social justice, or celebrating Black cultural contributions. For instance: Honoring the resilience and achievements of the Black community—Black Lives Matter.
    - **LGBTQ:** Messages in this classification aim to uplift the LGBTQ community by promoting equality, celebrating diversity, or supporting their rights and needs of representation. An example would be: Love knows no boundaries—proud to stand with the LGBTQ community this Pride Month.
    - **Religion:** Support in this category relates to specific religions or the general rights of religious communities, often emphasizing respect, solidarity, or advocacy against discrimination. For example: We stand with people of all faiths to ensure freedom of religion for everyone.

– **Women:** Texts highlighting gender equality, recognizing women's achievements, or supporting feminist causes fall here. An example could be: "Here's to empowering women to lead, inspire, and break barriers every day.

– **Other:** This group covers expressions of support directed at communities not explicitly listed in the categories. These could be niche groups or professional communities such as healthcare workers, teachers, or activists. For instance: Heartfelt gratitude to all the frontline workers saving lives during the pandemic.

*4.4. Annotation procedure*

The annotation process provided detailed guidelines and sample data to the three selected annotators to efficiently create the proposed dataset. In this sense, the annotators followed a structured process, as illustrated in Figure 1. This process begins by determining whether the comments expressed support in terms of concern, advocacy, happiness, or care. If any was detected, then it is labelled as Support. The second level of analysis consists in distinguishing whether the identified support content was directed to an individual or a group. In the case of a group, the annotators also specified the affiliation of the group, such as Nation, Religion, Black Community, Women, LGBTQ or Other. On the contrary, if the comment did not show support, the annotators marked it as Non-Support. This process ensures the quality of the labels and data set.

*4.5. Inter-annotator agreement*

Inter-annotator agreement (IAA) assesses how much annotators agree, factoring in chance agreement. Cohen's Kappa Coefficient scores of 0.84 for subtask 1, 0.78 for subtask 2, and 0.62 for subtask 3 demonstrate the robustness of the datasets, reflecting the rigorous annotation process.

*4.6. Statistics of the dataset*

Table 1 presents the dataset statistics, highlighting the variation for each subtask. In subtask 1, the proportion of **Non Support** samples greatly exceeds the proportion of **Support** samples, suggesting that there are few people who are willing to show support on YouTube comments. Subtask 2 shows that there is a greater propensity to support groups rather than individuals. Subtask 3 reveals which trend topics have generated the most supportive comments. It can be seen that LGBTQ people have recieved more support than other topics, suggesting that this issue has triggered many supportive comments in the Spanish-speaking community; while the level of support for LGBTQ is considerably lower in other topics, this behaviour suggests that these topics depend on various factors such as social impact generated, affinity, and empathy. This behaviour can still be studied in depth to determine the origin of support.

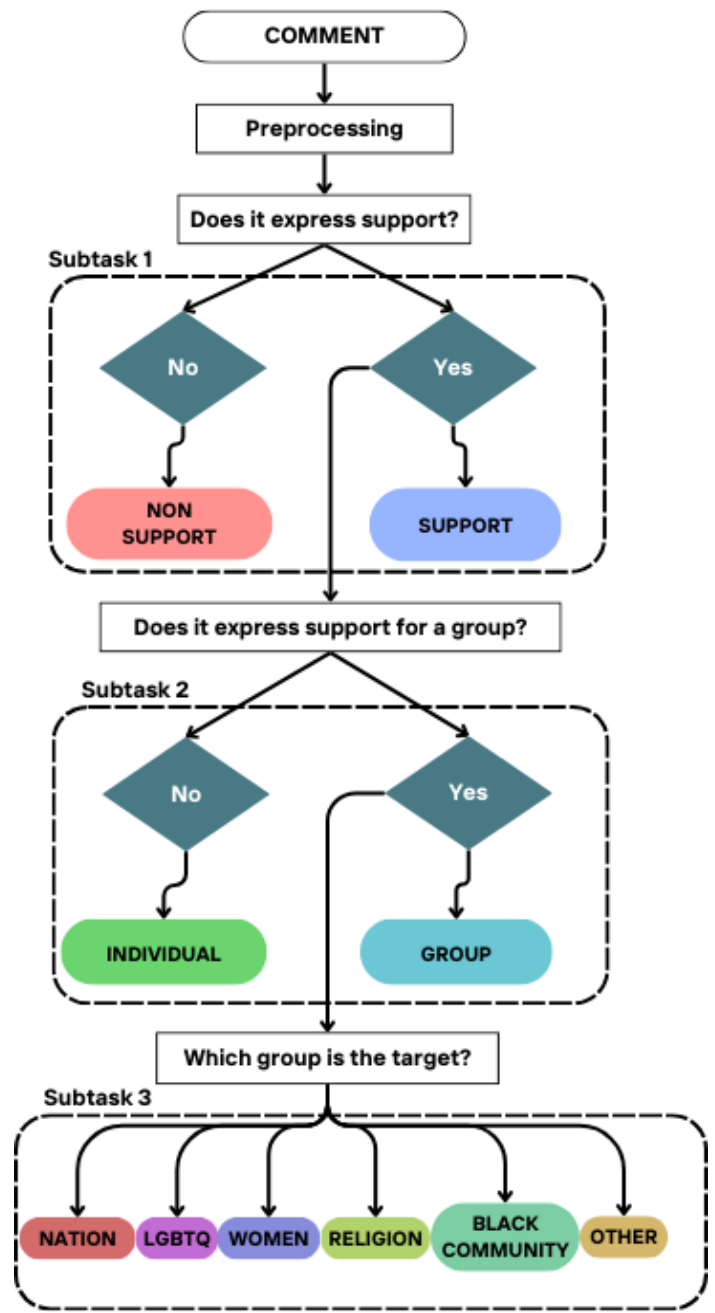

Figure 1: Overview of annotation procedure

| Tasks | Category | Number of samples |
|---|---|---|
| **Subtask1** | Social Support | 679 |
| | Non Support | 2510 |
| **Subtask2** | Individual | 171 |
| | Group | 508 |
| **Subtask3** | Nation | 35 |
| | Other | 101 |
| | LGBTQ | 245 |
| | Black Community | 16 |
| | Women | 41 |
| | Religion | 70 |

Table 1: Statistics of the dataset

## 5. Model training and Evalution

Before training, the dataset underwent preprocessing, including text cleaning, tokenization, and normalization, to improve model performance. To ensure a fair evaluation, we applied k-fold cross-validation, allowing for a robust assessment of both traditional and advanced machine learning models.

Traditional models were trained using standard feature extraction techniques, while advanced models leveraged deep learning architectures and pre-trained transformers to enhance text representation. We computed precision, recall, and F1-score using both weighted and macro averaging. Macro F1 treats all classes equally by averaging their individual F1-scores, making it suitable for balanced datasets. In contrast, the Weighted F1-score addresses class imbalance by assigning more weight to classes with more instances, meaning that larger classes have a greater influence on the final score. However, we selected the macro F1-score as the primary evaluation metric because it offers a fair assessment of performance across all classes, giving equal importance to each class regardless of its frequency.

The formulas for Precision, Recall, and F1-score are as follows (Derczynski, 2016):

$$\text{Precision} = \frac{TP}{TP + FP} \quad (1)$$

$$\text{Recall} = \frac{TP}{TP + FN} \quad (2)$$

$$\text{F1-score} = 2 \times \frac{\text{Precision} \times \text{Recall}}{\text{Precision} + \text{Recall}} \quad (3)$$

**Macro F1-score** calculates the F1-score for each class separately and then averages them:

$$\text{Macro-F1} = \frac{1}{N} \sum_{i=1}^{N} \text{F1-score}_i \quad (4)$$

where *N* is the number of classes.

**Weighted F1-score** takes class imbalance into account by weighting each class's F1-score based on its support:

$$\text{Weighted-F1} = \sum_{i=1}^{N} \frac{\text{support}_i}{\text{total instances}} \times \text{F1-score}_i \quad (5)$$

### 5.1. K-Fold Cross-Validation

K-fold cross-validation (CV) is a technique used to evaluate machine learning models (MLMs) by dividing the dataset into k folds. In each iteration, one fold serves as the test data, while the remaining folds are used for training. This process is repeated until the entire dataset has been tested. The results are typically averaged to calculate the mean score of the MLM. In this study, we selected k=5 and compared the classification performance of both advanced and traditional machine learning models using their respective classification reports (Nti et al., 2021).

### 5.2. Traditional machine learning models

For our classification task, we selected five traditional machine learning models to ensure a comprehensive evaluation of different algorithms. These models include Logistic Regression (LR), Support Vector Machine (SVM) with both radial basis function (RBF) and linear kernels, XGBoost, and Random Forest. These models are well-established in the field of text classification, known for their effectiveness in handling a variety of tasks (Ahani et al., 2024). To represent the text data effectively for these models, we used TF-IDF (Term Frequency-Inverse Document Frequency) as the feature extraction technique. TF-IDF helps to capture the importance of words in relation to the entire dataset, by weighing terms based on their frequency within a document and across the corpus (Roelleke and Wang, 2008).

### 5.3. Deep learning

For our text classification task, we utilized deep learning models with different word embeddings to enhance performance. We implemented **CNN** and **BiLSTM** architectures with both **GloVe** and **FastText** embeddings, allowing the models to capture contextual meaning and semantic relationships effectively (Kolesnikova et al., 2025).

### 5.4. Transformers

Transformers are deep learning models widely used in NLP tasks like translation, summarization, and text generation. They employ self-attention mechanisms to capture long-range dependencies, allowing efficient text processing.

For our text classification task, we utilized the Hugging Face **Transformers** library, leveraging pre-trained models such as **XLM-RoBERTa**, **RoBERTuito**, **BERT**, and **DistilBERT** to enhance accuracy and efficiency (Kolesnikova et al., 2025).

### 5.5. GPT

GPT-4o is a powerful Transformer-based model pre-trained on vast text data. It excels in NLP tasks like text classification, sentiment analysis, and text generation.

**Pre-training:** The model learns language patterns by predicting words in large datasets without labeled data. **Fine-tuning:** It is further trained on specific datasets, like our support dataset, to enhance classification accuracy. **Classification:** GPT-4o analyzes text contextually, assigning relevant labels based on learned patterns. **Contextual Understanding:** Using attention mechanisms, it ensures precise and relevant classifications (Imamguluyev, 2023).

## 5.6. Balanced Data Set

Initially, the dataset was divided into training and test sets, with 20% allocated for testing. To address class imbalance in the training data, we applied an oversampling technique using the GPT-4-O model. This model generated paraphrased versions of comments to augment the underrepresented classes, ensuring a more balanced distribution. Detailed statistics of the dataset can be found in Table 2.

| Tasks | Category | Number of Sample Train Set | Number of Sample Test Set |
|---|---|---|---|
| Subtask1 | Social Support | 2017 | 146 |
| | Non Support | 2017 | 495 |
| Subtask2 | Individual | 397 | 33 |
| | Group | 397 | 124 |
| Subtask3 | Nation | 192 | 9 |
| | Other | 192 | 29 |
| | LGBTQ | 192 | 62 |
| | Black Community | 192 | 3 |
| | Women | 192 | 8 |
| | Religion | 192 | 13 |

Table 2: Distribution of training and test samples across different tasks and categories after applying oversampling

## 6. Results

### 6.1. Traditional machine learning

Table 3 presents the performance of various traditional machine learning models for Subtask 1. The results are reported in terms of weighted and macro scores for precision, recall, and F1-score, along with accuracy.

Among the models, the linear SVM achieved the highest weighted F1-score of 0.8445 and the best accuracy of 86.04%. It also demonstrated strong weighted precision and recall, highlighting its overall reliability in this task.

The XGBoost and Random Forest classifiers produced comparable results, with XGBoost slightly outperforming Random Forest in weighted recall (0.8464 vs. 0.8511) but achieving a marginally lower macro F1-score. Both models performed better than Logistic Regression (LR) and SVM with RBF kernels in terms of macro-level precision and recall.

Interestingly, while Logistic Regression and SVM (RBF) delivered similar weighted F1-scores (0.8298 and 0.8311, respectively), their macro scores were slightly lower, indicating potential challenges in handling imbalanced class distributions compared to other models.

Overall, XGBoost emerges as the most effective model, demonstrating superior performance with a Macro F1 score of 0.7577, highlighting its robustness for the classification task in Subtask 1

| Model | Weighted Scores | | | Macro Scores | | | Accuracy |
|---|---|---|---|---|---|---|---|
| | Precision | Recall | F1-score | Precision | Recall | F1-score | |
| Logistic Regression | 0.8575 | 0.8541 | 0.8298 | 0.8637 | 0.6766 | 0.7153 | 0.8541 |
| SVM (linear) | 0.8567 | 0.8604 | 0.8445 | 0.8418 | 0.7100 | 0.7467 | 0.8604 |
| SVM (RBF) | 0.8574 | 0.8548 | 0.8311 | 0.8621 | 0.6794 | 0.7184 | 0.8548 |
| XGBoost | 0.8374 | 0.8464 | 0.8386 | 0.7908 | 0.7369 | **0.7577** | 0.8464 |
| Random Forest | 0.8425 | 0.8511 | 0.8392 | 0.8115 | 0.7250 | 0.7539 | 0.8511 |

Table 3: Traditional machine learning models performance in Subtask 1

Table 4 summarizes the performance of various traditional machine learning models for Subtask 2. Metrics are reported as weighted and macro scores for precision, recall, and F1-score, along with overall accuracy.

The linear SVM achieved the best overall performance, with a weighted F1-score of 0.8432, a macro F1-score of 0.7763, and an accuracy of 85.42%. Its consistent performance across all metrics highlights its effectiveness in this classification task. Similarly, XGBoost closely followed, achieving a weighted F1-score of 0.8379, a macro F1-score of 0.7831, and an accuracy of 84.56%. These results suggest XGBoost's ability to balance performance between majority and minority classes effectively.

While Logistic Regression (LR) and SVM with RBF kernels demonstrated decent weighted precision and recall, their macro F1-scores (0.6452 and 0.6870, respectively) indicate that these models struggled with imbalanced class distributions. Random Forest, while achieving moderate accuracy (80.88%) and weighted F1-score (0.7886), exhibited lower macro scores, suggesting that its predictions were skewed towards majority classes.

Overall, XGBoost proved to be the best-performing model in Subtask 2, achieving a macro F1 score of 0.7831.

| Model | Weighted Scores | | | Macro Scores | | | Accuracy |
|---|---|---|---|---|---|---|---|
| | Precision | Recall | F1-score | Precision | Recall | F1-score | |
| Logistic Regression | 0.8290 | 0.8086 | 0.7659 | 0.8528 | 0.6286 | 0.6452 | 0.8086 |
| SVM (linear) | 0.8513 | 0.8542 | 0.8432 | 0.8363 | 0.7503 | 0.7763 | 0.8542 |
| SVM (RBF) | 0.8330 | 0.8203 | 0.7892 | 0.8433 | 0.6610 | 0.6870 | 0.8203 |
| XGBoost | 0.8400 | 0.8456 | 0.8379 | 0.8201 | 0.7617 | **0.7831** | 0.8456 |
| Random Forest | 0.8007 | 0.8088 | 0.7886 | 0.7845 | 0.6833 | 0.7077 | 0.8088 |

Table 4: Traditional machine learning models performance in Subtask 2

Table 5 displays the performance of traditional machine learning models for Subtask 3, evaluated using weighted and macro scores for precision, recall, and F1-score, along with overall accuracy.

The Random Forest model achieved the highest weighted precision (0.8152) and accuracy (77.45%), along with a strong weighted F1-score of 0.7554. Its macro scores (precision: 0.8119, recall: 0.7176, F1-score: 0.7096) indicate robust performance across all classes, making it the most effective model for Subtask 3.

The linear SVM demonstrated solid performance with a weighted F1-score of 0.7752 and a macro F1-score of 0.6309, indicating its ability to maintain balanced predictions. While its macro-level scores were slightly lower than Random Forest, it still outperformed other models.

XGBoost followed closely with a weighted F1-score of 0.7155 and a macro F1-score of 0.6964. Its macro precision (0.7708) and recall (0.6889) suggest it managed class imbalance better than some models, but its performance lagged behind Random Forest and linear SVM.

Both Logistic Regression (LR) and SVM with RBF kernel showed lower macro F1-scores (0.5234 and 0.5023, respectively), reflecting challenges in addressing class imbalance. While their weighted scores and accuracies were competitive, their macro-level metrics reveal limitations in handling minority classes effectively.

Overall, Random Forest achieved the highest macro F1 score in Subtask 3, while linear SVM and XGBoost also performed well.

| Model | Weighted Scores | | | Macro Scores | | | |
|---|---|---|---|---|---|---|---|
| | Precision | Recall | F1-score | Precision | Recall | F1-score | Accuracy |
| Logistic Regression | 0.7376 | 0.7304 | 0.6909 | 0.6798 | 0.4833 | 0.5234 | 0.7304 |
| SVM (linear) | 0.8034 | 0.7993 | 0.7752 | 0.7553 | 0.5927 | 0.6309 | 0.7993 |
| SVM (RBF) | 0.7332 | 0.7206 | 0.6793 | 0.6764 | 0.4602 | 0.5023 | 0.7206 |
| XGBoost | 0.7716 | 0.7353 | 0.7155 | 0.7708 | 0.6889 | 0.6964 | 0.7353 |
| Random Forest | 0.8152 | 0.7745 | 0.7554 | 0.8119 | 0.7176 | **0.7096** | 0.7745 |

Table 5: Traditional machine learning models performance in Subtask 3

*6.2. Deep learning*

Table 6 presents the performance of various deep learning models for Subtask 1. These models were evaluated using weighted and macro precision, recall, and F1-scores, as well as overall accuracy.

The BiLSTM (GloVe embeddings) achieved the highest weighted F1-score of 0.8273 and an accuracy of 83.62%, demonstrating its effectiveness in capturing contextual relationships in text. Its macro F1-score of 0.7314 further highlights its ability to balance predictions across all classes, outperforming the other models in this metric.

The BiLSTM (FastText embeddings) showed comparable performance, with a weighted F1-score of 0.8267 and an accuracy of 83.72%. While slightly behind the GloVe-based BiLSTM in macro F1-score (0.7281 vs. 0.7314), it demonstrated strong consistency across metrics, making it a reliable alternative.

Among the convolutional models, the CNN (FastText embeddings) outperformed its GloVe counterpart, achieving a weighted F1-score of 0.8203 and an accuracy of 83.91%. Its macro precision (0.7799) and recall (0.6936) suggest it effectively leverages word-level representations provided by FastText.

The CNN (GloVe embeddings), while achieving the lowest macro F1-score (0.5964) and weighted F1-score (0.7643), still demonstrated reasonable accuracy (80.69%). This indicates that while CNNs may excel at identifying local patterns, they may not capture global dependencies as effectively as BiLSTM models.

Overall, the BiLSTM models, particularly with GloVe embeddings, emerged as the most effective for Subtask 1, emphasizing the importance of sequential modeling and rich word embeddings in text classification tasks.

| Model | Weighted Scores | | | Macro Scores | | | |
|---|---|---|---|---|---|---|---|
| | Precision | Recall | F1-score | Precision | Recall | F1-score | Accuracy |
| CNN (glove) | 0.7812 | 0.8069 | 0.7643 | 0.7306 | 0.5850 | 0.5964 | 0.8069 |
| BiLSTM (glove) | 0.8327 | 0.8362 | 0.8273 | 0.7554 | 0.7350 | **0.7314** | 0.8362 |
| CNN (fasttext) | 0.8292 | 0.8391 | 0.8203 | 0.7799 | 0.6936 | 0.7085 | 0.8391 |
| BiLSTM (fasttext) | 0.8329 | 0.8372 | 0.8267 | 0.7667 | 0.7220 | 0.7281 | 0.8372 |

Table 6: Deep Learning models performance in Subtask 1

Table 7 illustrates the performance of various deep learning models in Subtask 2. The results are reported in terms of weighted and macro precision, recall, and F1-scores, alongside accuracy.

The CNN (FastText embeddings) emerged as the best-performing model, achieving a weighted F1-score of 0.7736 and an accuracy of 80.11%. Its macro precision (0.7916) further highlights its ability to maintain a high degree of class distinction, although its macro F1-score (0.6726) indicates room for improvement in balancing performance across all classes.

The CNN (GloVe embeddings) showed comparable performance with a weighted F1-score of 0.7708 and an accuracy of 79.67%. While it slightly lagged behind its FastText counterpart in weighted scores, its macro precision (0.7350) and recall (0.6803) demonstrate reasonable performance in handling class imbalances.

In contrast, the BiLSTM models yielded lower weighted F1-scores compared to CNNs. The BiLSTM (GloVe embeddings) achieved a weighted F1-score of 0.6701 and the highest macro F1-score of 0.7548, demonstrating better balance in predictions across classes despite a lower overall accuracy (76.27%). Similarly, the BiLSTM (FastText embeddings) attained a weighted F1-score of 0.7477 and an accuracy of 76.72%, showcasing its reliability in sequential tasks but slightly underperforming in handling class variability.

Overall, CNN models excelled in achieving higher accuracy and weighted scores, indicating their proficiency in leveraging pre-trained embeddings for overall task performance. However, the superior macro F1-scores of BiLSTM models highlight their advantage in addressing class-level disparities, especially with GloVe embeddings.

| Model | Weighted Scores | | | Macro Scores | | | |
|---|---|---|---|---|---|---|---|
| | Precision | Recall | F1-score | Precision | Recall | F1-score | Accuracy |
| CNN (glove) | 0.7883 | 0.7967 | 0.7708 | 0.7350 | 0.6803 | 0.6715 | 0.7967 |
| BiLSTM (glove) | 0.6913 | 0.6812 | 0.6701 | 0.7691 | 0.7627 | **0.7548** | 0.7627 |
| CNN (fasttext) | 0.8059 | 0.8011 | 0.7736 | 0.7916 | 0.6608 | 0.6726 | 0.8011 |
| BiLSTM (fasttext) | 0.7897 | 0.7672 | 0.7477 | 0.7446 | 0.6627 | 0.6521 | 0.7672 |

Table 7: Deep learning models performance in Subtask 2

Table 8 presents the performance metrics for deep learning models applied to Subtask 3, showcasing their weighted and macro scores for precision, recall, F1, and overall accuracy.

Among the models evaluated, the CNN (GloVe embeddings) stands out with the highest weighted F1-score of 0.7960 and an accuracy of 82.07%. Its macro F1-score of 0.6489 indicates a relatively balanced performance across different classes compared to other models, making it the most effective in handling this subtask.

The BiLSTM (GloVe embeddings) achieved a weighted F1-score of 0.7420 and an accuracy of 75.38%. While it trails the CNN (GloVe) in weighted scores, its macro F1-score of 0.6063 reflects its ability to perform consistently across multiple classes, albeit with some limitations.

In contrast, the models using FastText embeddings performed less competitively. The CNN (FastText) scored a weighted F1 of 0.6489 and an accuracy of 69.07%, while the BiLSTM (FastText) achieved slightly better weighted F1 and accuracy values of 0.6549 and 67.11%, respectively. Both models, however, showed limited ability to balance class-specific performance, as reflected in their lower macro F1-scores (0.4595 for CNN and 0.4839 for BiLSTM).

These results highlight the effectiveness of GloVe embeddings over FastText embeddings in this task, with CNN architectures leveraging these embeddings most effectively. While BiLSTM models showed relatively consistent performance, their lower overall scores suggest that they may not fully capture the intricacies of the task compared to CNNs.

| Model | Weighted Scores | | | Macro Scores | | | |
|---|---|---|---|---|---|---|---|
| | Precision | Recall | F1-score | Precision | Recall | F1-score | Accuracy |
| CNN (glove) | 0.7917 | 0.8207 | 0.7960 | 0.6951 | 0.6446 | **0.6489** | 0.8207 |
| BiLSTM (glove) | 0.7453 | 0.7538 | 0.7420 | 0.6289 | 0.6059 | 0.6063 | 0.7538 |
| CNN (fasttext) | 0.6370 | 0.6907 | 0.6489 | 0.4891 | 0.4659 | 0.4595 | 0.6907 |
| BiLSTM (fasttext) | 0.6672 | 0.6711 | 0.6549 | 0.5097 | 0.4992 | 0.4839 | 0.6711 |

Table 8: Deep learning models performance in Subtask 3

*6.3. Transformers*

Table 9 shows the performance of various transformer-based models in Subtask 1, providing their weighted and macro scores for precision, recall, F1-score, and accuracy.

The model pysentimiento/robertuito-sentiment-analysis stands out with the highest performance in this subtask. It achieved a weighted F1-score of 0.8673 and an accuracy of 87.27%, significantly outpacing the others. Its macro F1-score of 0.7956 further highlights its ability to maintain strong performance across all classes, reflecting its robust handling of the task.

Another strong contender is papluca/xlm-roberta-base-language-detection, which achieved a weighted F1-score of 0.8386 and an accuracy of 84.60%. This model is also highly effective in balancing precision and recall, with macro precision and recall values of 0.7837 and 0.7482, respectively. Its performance is highly competitive, though slightly behind the robertuito-sentiment-analysis model.

The nlptown/bert-base-multilingual-uncased-sentiment model also shows strong results, achieving a weighted F1-score of 0.8416 and an accuracy of 85.29%. Its macro F1-score is 0.7502, demonstrating a solid balance between precision and recall, positioning it as another strong performer.

The lxyuan/distilbert-base-multilingual-cased-sentiments-student model comes next with a weighted F1-score of 0.8329 and accuracy of 84.35%. While it trails the other transformers in terms of F1-score, it maintains a consistent performance across the task with a macro F1-score of 0.7372.

Lastly, papluca/xlm-roberta-base-language-detection (second instance) shows a very similar performance to its first instance, achieving a weighted F1-score of 0.8415 and an accuracy of 84.95%. This indicates that the XLM-RoBERTa model consistently delivers competitive results across different runs.

In summary, pysentimiento/robertuito-sentiment-analysis leads in terms of both weighted and macro scores, making it the top-performing transformer model for Subtask 1.

| Model | Weighted Scores | | | Macro Scores | | | |
|---|---|---|---|---|---|---|---|
| | Precision | Recall | F1-score | Precision | Recall | F1-score | Accuracy |
| papluca/xlm-roberta-base-language-detection | 0.8476 | 0.8460 | 0.8386 | 0.7837 | 0.7482 | 0.7506 | 0.8460 |
| pysentimiento/robertuito-sentiment-analysis | 0.8764 | 0.8727 | 0.8673 | 0.8245 | 0.7942 | **0.7956** | 0.8727 |
| nlptown/bert-base-multilingual-uncased-sentiment | 0.8528 | 0.8529 | 0.8416 | 0.7949 | 0.7498 | 0.7502 | 0.8529 |
| lxyuan/distilbert-base-multilingual-cased-sentiments-student | 0.8393 | 0.8435 | 0.8329 | 0.7774 | 0.7304 | 0.7372 | 0.8435 |
| papluca/xlm-roberta-base-language-detection | 0.8528 | 0.8495 | 0.8415 | 0.7920 | 0.7541 | 0.7547 | 0.8495 |

Table 9: Transformers models performance in Subtask 1

Table 10 displays the performance of transformer-based models on Subtask 2. The table includes weighted and macro scores for precision, recall, F1-score, and accuracy for each model.

The pysentimiento/robertuito-sentiment-analysis model outperforms the others, achieving a weighted F1-score of 0.8598 and an accuracy of 86.26%. Its macro F1-score is 0.8143, showing a strong balance between precision and recall across the task. This model's consistent high performance across all metrics positions it as the top performer for Subtask 2.

The papluca/xlm-roberta-base-language-detection model shows solid results, with a weighted F1-score of 0.7356 and an accuracy of 72.83%. Despite being a bit behind in F1-score and accuracy, this model maintains a good balance between precision and recall, with a macro F1-score of 0.6739.

The nlptown/bert-base-multilingual-uncased-sentiment model demonstrates strong results as well, achieving a weighted F1-score of 0.7894 and an accuracy of 79.03%. Its macro F1-score is 0.7278, which suggests its ability to handle the task reasonably well with a solid balance between precision and recall.

The lxyuan/distilbert-base-multilingual-cased-sentiments-student model performs well with a weighted F1-score of 0.7226 and an accuracy of 72.84%. Although it lags behind the other transformers, it still shows respectable results with a macro F1-score of 0.6435, making it a reliable option for Subtask 2.

The second instance of papluca/xlm-roberta-base-language-detection exhibits a stronger performance than the first, achieving a weighted F1-score of 0.7508 and an accuracy of 75.65%. Its macro F1-score of 0.6814 further confirms its competitive performance.

In summary, pysentimiento/robertuito-sentiment-analysis remains the top-performing transformer model in Subtask 2, leading in terms of F1-score and accuracy.

| Model | Weighted Scores | | | Macro Scores | | | |
|---|---|---|---|---|---|---|---|
| | Precision | Recall | F1-score | Precision | Recall | F1-score | Accuracy |
| papluca/xlm-roberta-base-language-detection | 0.7793 | 0.7283 | 0.7356 | 0.6884 | 0.7034 | 0.6739 | 0.7283 |
| pysentimiento/robertuito-sentiment-analysis | 0.8875 | 0.8626 | 0.8598 | 0.8479 | 0.8302 | **0.8143** | 0.8626 |
| nlptown/bert-base-multilingual-uncased-sentiment | 0.8184 | 0.7903 | 0.7894 | 0.7495 | 0.7448 | 0.7278 | 0.7903 |
| lxyuan/distilbert-base-multilingual-cased-sentiments-student | 0.7681 | 0.7284 | 0.7226 | 0.6857 | 0.6644 | 0.6435 | 0.7284 |
| papluca/xlm-roberta-base-language-detection | 0.7929 | 0.7565 | 0.7508 | 0.7163 | 0.7026 | 0.6814 | 0.7565 |

Table 10: Transformers models performance in Subtask 2

Table 11 presents the performance of transformer-based models in Subtask 3. The table shows weighted and macro scores for precision, recall, F1-score, and accuracy for each model.

pysentimiento/robertuito-sentiment-analysis achieves the highest performance, with a weighted F1-score of 0.8715 and an accuracy of 87.98%. The model also excels with a macro F1-score of 0.8177, showcasing a strong balance between precision and recall across all classes. This model stands out as the top performer for Subtask 3.

The papluca/xlm-roberta-base-language-detection model demonstrates solid results, with a weighted F1-score of 0.8186 and an accuracy of 82.67%. It also achieves a good macro F1-score of 0.7540, indicating its effectiveness in handling the task.

nlptown/bert-base-multilingual-uncased-sentiment follows closely, with a weighted F1-score of 0.8546 and an accuracy of 87.20%. Its macro F1-score of 0.8088 suggests a strong performance with balanced precision and recall, though it slightly lags behind the top performers in F1-score.

The lxyuan/distilbert-base-multilingual-cased-sentiments-student model performs well, achieving a weighted F1-score of 0.8271 and an accuracy of 84.45%. Its macro F1-score of 0.7735 indicates a solid performance, though slightly lower than the top models.

Another instance of papluca/xlm-roberta-base-language-detection shows strong performance, with a weighted F1-score of 0.8268 and an accuracy of 83.64%. This model maintains consistency across different tasks, performing well in Subtask 3 as well.

In summary, pysentimiento/robertuito-sentiment-analysis is the best-performing model in Subtask 3, excelling in both weighted and macro F1-scores as well as accuracy.

| Model | Weighted Scores | | | Macro Scores | | | Accuracy |
|---|---|---|---|---|---|---|---|
| | Precision | Recall | F1-score | Precision | Recall | F1-score | |
| papluca/xlm-roberta-base-language-detection | 0.8520 | 0.8267 | 0.8186 | 0.8097 | 0.7626 | 0.7540 | 0.8267 |
| pysentimiento/robertuito-sentiment-analysis | 0.8852 | 0.8798 | 0.8715 | 0.8486 | 0.8219 | **0.8177** | 0.8798 |
| nlptown/bert-base-multilingual-uncased-sentiment | 0.8729 | 0.8720 | 0.8546 | 0.8544 | 0.8171 | 0.8088 | 0.8720 |
| lxyuan/distilbert-base-multilingual-cased-sentiments-student | 0.8650 | 0.8445 | 0.8271 | 0.8363 | 0.7815 | 0.7735 | 0.8445 |
| papluca/xlm-roberta-base-language-detection | 0.8429 | 0.8364 | 0.8268 | 0.7810 | 0.7620 | 0.7502 | 0.8364 |

Table 11: Transformers models performance in Subtask 3

### 6.4. GPT

Table 12 shows the performance of the GPT4-o model across three subtasks, providing both weighted and macro scores for precision, recall, F1-score, and accuracy.

Subtask 1: The GPT4-o model performs exceptionally well with a weighted F1-score of 0.9006 and an accuracy of 89.97%. The model demonstrates a solid balance between precision and recall with macro F1-scores of 0.8531, making it a strong performer for this subtask.

Subtask 2: For Subtask 2, GPT4-o achieves a weighted F1-score of 0.6322 and accuracy of 61.27%. The macro F1-score (0.6051) indicates some challenges in this task, particularly in terms of recall, where it drops to 0.6127. This suggests that GPT4-o may have more difficulty distinguishing certain classes in this subtask.

Subtask 3: The model performs strongly in Subtask 3, with a weighted F1-score of 0.8472 and an accuracy of 84.45%. The macro F1-score of 0.7903 reflects a good balance of precision and recall, showing that the model handles this task well.

In summary, GPT4-o excels in Subtasks 1 and 3, with excellent precision, recall, and accuracy. However, in Subtask 2, the model's performance drops, particularly in recall. Nonetheless, it remains a strong model overall across the subtasks.

### 6.5. Balanced Data set

Table 13 presents the performance metrics of the robertuito-sentiment-analysis model across three tasks, using a balanced dataset. The evaluation includes weighted and macro scores for precision, recall, F1-score, and overall accuracy.

| Task | Weighted Scores | | | Macro Scores | | | Accuracy |
|---|---|---|---|---|---|---|---|
| | Precision | Recall | F1-score | Precision | Recall | F1-score | |
| SubTask 1 | 0.9018 | 0.8997 | 0.9006 | 0.8468 | 0.8600 | **0.8531** | 0.8997 |
| SubTask 2 | 0.8179 | 0.6127 | 0.6322 | 0.6756 | 0.7217 | **0.6051** | 0.6127 |
| SubTask 3 | 0.8969 | 0.8445 | 0.8472 | 0.8301 | 0.8229 | **0.7903** | 0.8445 |

Table 12: GPT4-o performance

Task 1: The model performs well with a weighted F1-score of 0.89, showing good precision (0.89) and recall (0.89). The macro F1-score is 0.84, indicating a slightly lower performance on individual classes compared to the overall weighted score, but still reflecting strong performance overall. The model achieves an accuracy of 89

Task 2: The model achieves its best performance in this task with a weighted F1-score of 0.92, driven by excellent precision (0.93) and recall (0.92). The macro F1-score is 0.89, which suggests that the model handles individual classes well across the task. The accuracy reaches 92%, indicating strong overall effectiveness.

Task 3: The model's performance is still good but slightly lower than in Tasks 1 and 2, with a weighted F1-score of 0.87. Precision is 0.87, while recall is 0.88, reflecting a solid balance. The macro F1-score drops to 0.72, indicating that the model might struggle slightly with certain classes. However, the model achieves an accuracy of 88%.

In summary, the robertuito-sentiment-analysis model performs consistently well across all tasks with strong precision and recall, achieving the highest accuracy in Task 2 (92%). Task 3 shows a slightly lower macro performance, but the model still maintains high accuracy in all subtasks.

| Task | Model | Weighted | | | Macro | | | Accuracy |
|---|---|---|---|---|---|---|---|---|
| | | Precision | Recall | F1-Score | Precision | Recall | F1-Score | |
| SubTask 1 | robertuito-sentiment-analysis | 0.8873 | 0.8830 | 0.8847 | 0.8295 | 0.8494 | **0.8387** | 0.8830 |
| SubTask 2 | | 0.9261 | 0.9178 | 0.9201 | 0.8706 | 0.9147 | **0.8894** | 0.9178 |
| SubTask 3 | | 0.9018 | 0.8850 | 0.8842 | 0.8969 | 0.8173 | **0.8361** | 0.8850 |

Table 13: Performance metrics for the robertuito-sentiment-analysis model on a balanced dataset.

### 6.6. Best performance

According to the results of all models evaluated in this Study, the models presented in Table 14 demonstrate the best performance compared to others, as measured by macro F1 scores across different tasks. Specifically, the balanced dataset approach achieved the highest F1-scores for SubTask 2 and SubTask 3, while the GPT4-o model showed strong performance for SubTask 1. It is important to note that the balanced dataset experiment was also conducted with the obertuito-sentiment-analysis model, further enhancing the overall performance.

| Task | Model | Weighted | | | Macro | | | Accuracy |
|---|---|---|---|---|---|---|---|---|
| | | Precision | Recall | F1-score | Precision | Recall | F1-score | |
| SubTask 1 | GPT4-o | 0.9018 | 0.8997 | 0.9006 | 0.8468 | 0.8600 | 0.8531 | 0.8997 |
| SubTask 2 | Balanced Dataset | 0.9261 | 0.9178 | 0.9201 | 0.8706 | 0.9147 | 0.8894 | 0.9178 |
| SubTask 3 | Balanced Dataset | 0.9018 | 0.8850 | 0.8842 | 0.8969 | 0.8173 | 0.8361 | 0.8850 |

Table 14: Best performance metrics across all tasks based on weighted and macro F1-scores.

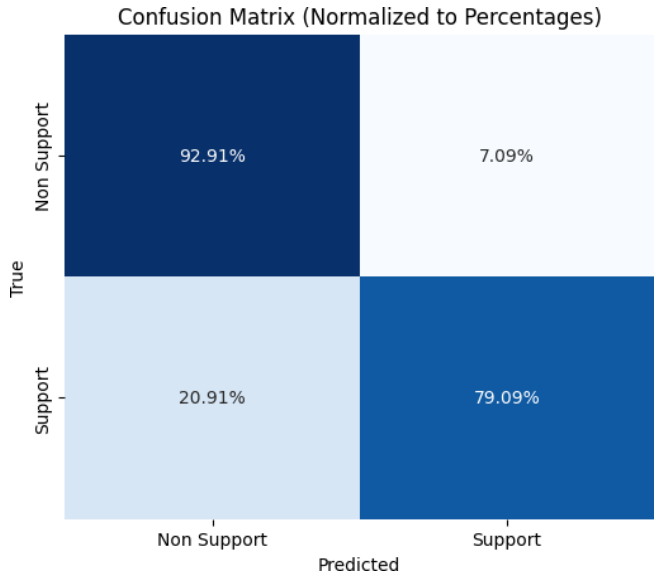

Figure 2: Confusion Matrix for SubTask 1

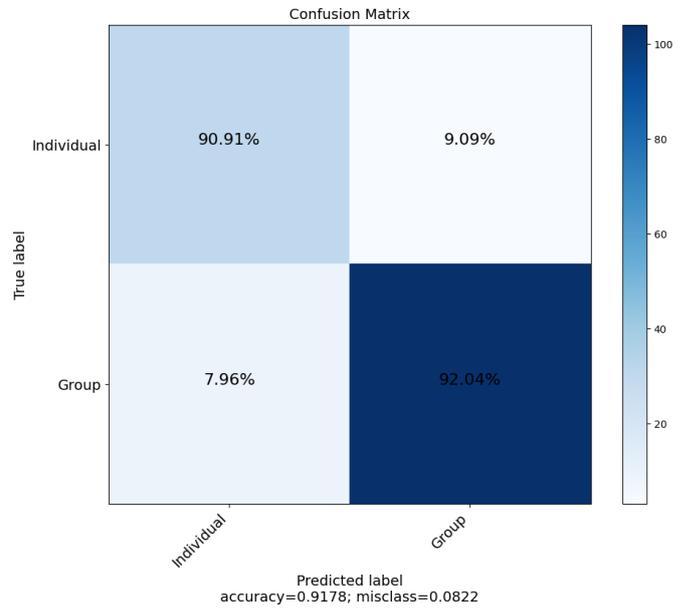

Figure 3: Confusion Matrix for SubTask 2

## 7. Error analysis

The table (15) presents the precision, recall, and F1-score for different labels across various tasks. For SubTask 1, the GPT4-O model achieved strong performance, particularly for the NSS label with an F1-score of 0.9358, while SS had a slightly lower F1-score of 0.7704. In SubTask 2, the balanced dataset showed excellent performance, with Group achieving an F1-score of 0.9455 and Individual scoring 0.8333. For SubTask 3, the balanced dataset also produced high F1-scores across different labels, including LGBTQ (0.9608) and Women (0.8421). Notably, the Nation label had an F1-score of 0.7143, reflecting some challenges. To better understand the model's performance, confusion matrix figures 2, 3, and 4 for the three tasks are provided, illustrating how the models performed across the different labels.

| Task | Model | Label | Precision | Recall | F1-score |
|---|---|---|---|---|---|
| SubTask 1 | GPT4-O | SS | 0.7510 | 0.7909 | 0.7704 |
|  |  | NSS | 0.9426 | 0.9291 | 0.9358 |
| SubTask 2 | Balance Data set | Individual | 0.7692 | 0.9091 | 0.8333 |
|  |  | Group | 0.9720 | 0.9204 | 0.9455 |
| SubTask 3 | Balance Data set | Other | 0.7576 | 0.8929 | 0.8197 |
|  |  | Nation | 1.0000 | 0.5556 | 0.7143 |
|  |  | LGBTQ | 0.9800 | 0.9423 | 0.9608 |
|  |  | Black Community | 1.0000 | 0.6667 | 0.8000 |
|  |  | Women | 0.7273 | 1.0000 | 0.8421 |
|  |  | Religion | 0.9167 | 0.8462 | 0.8800 |

Table 15: Precision, Recall, and F1-scores for Different Labels Across Tasks.

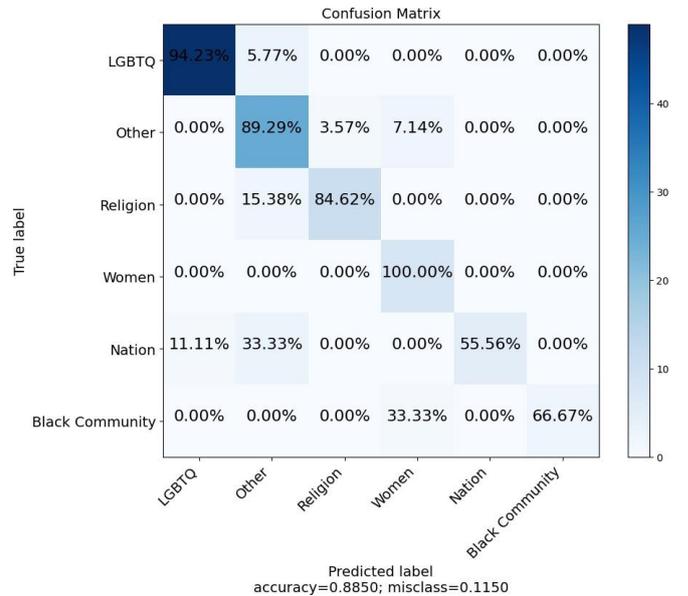

Figure 4: Confusion Matrix for SubTask 3

## 8. Discussion

Our findings underscore the effectiveness of NLP techniques in identifying social support within online discussions, showcasing their potential to foster a more positive digital environment. A comparison of machine learning, deep learning, and transformer-based models reveals that transformers, particularly pysentimiento/robertuito-sentiment-analysis, consistently outperform other approaches in Subtask2 and Subtask3 on a balanced dataset. For Task1, GPT-4o achieved the best performance on an unbalanced dataset. This suggests that contextual embeddings and large-scale pretraining play a critical role in enhancing classification accuracy for social support detection.

The use of GPT-4o to create a balanced dataset through paraphrasing led to significant improvements in model performance. By addressing class imbalances, the balanced dataset improved recall for minority classes. However, challenges remain in distinguishing between subtly supportive messages and neutral or ambiguous comments. Future research should focus on developing more refined annotation guidelines and advanced feature engineering techniques to improve classification granularity.

## 9. Conclusion and future work

This study presents the first comprehensive approach to detecting online social support in Spanish social media texts. Through a newly annotated dataset and extensive experimentation with various machine learning and deep learning models, we demonstrate the feasibility of automatic social support classification. Future research should focus on expanding the dataset with more diverse sources and refining the annotation process to capture a broader range of supportive expressions. Additionally, integrating real-time support detection into social media platforms could help identify and promote positive interactions, ultimately contributing to a more supportive online ecosystem. Another promising direction is the exploration of multimodal approaches that incorporate text, images, and video content to enhance the understanding of social support dynamics in online spaces.

## 10. Limitations

Despite the promising results, this study has several limitations. First, the dataset is limited to YouTube comments, which may not fully represent supportive communication across different social media platforms. Expanding the dataset to include platforms like Twitter, Facebook, and Reddit could improve the generalizability of the models. Second, the annotation process, while rigorous, is subject to human interpretation. The complexity and subtlety of social support expressions might lead to inconsistencies in labeling, which could impact model performance. Future studies should incorporate more annotators and develop clearer guidelines to reduce subjectivity. Third, while transformers demonstrated superior performance, their computational cost is significantly higher compared to traditional machine learning models. This limitation poses challenges for real-time deployment, particularly in resource-constrained environments. Future work should explore optimization techniques, such as model distillation, to reduce inference time without compromising accuracy. Finally, cultural and linguistic variations in social support expressions were not extensively analyzed in this study. A cross-cultural comparison of social support detection could provide deeper insights into how support is expressed differently across languages and communities.

## Declarations

*Funding*

The work was done with partial support from the Mexican Government through the grant A1-S-47854 of CONACYT, Mexico, grants 20241816, 20241819, and 20240951 of the Secretaría de Investigación y Posgrado of the Instituto Politécnico Nacional, Mexico. The authors thank the CONACYT for the computing resources brought to them through the Plataforma de Aprendizaje Profundo para Tecnologías del Lenguaje of the Laboratorio de Supercómputo of the INAOE, Mexico and acknowledge the support of Microsoft through the Microsoft Latin America PhD Award.

*Conflict of Interest*

I declare that the authors have no competing interests as defined by Nature Research, or other interests that might be perceived to influence the results and/or discussion reported in this paper.

*Availability of data and materials*

The dataset utilized in this study can be obtained upon request from the corresponding author. Please reach out to Moein Shahiki Tash at [mshahikit2022@cic.ipn.mx].